\def\year{2020}\relax
\documentclass[letterpaper]{article} 
\usepackage{aaai20}  
\usepackage{times}  
\usepackage{helvet} 
\usepackage{courier}  
\usepackage[hyphens]{url}  
\usepackage{graphicx} 
\usepackage{graphicx}  
\usepackage{xcolor}
\usepackage{amssymb}
\usepackage{amsmath}
\usepackage{diagbox}
\usepackage{tikz}
\usepackage{marvosym}

\usepackage{amsthm,amsmath,amssymb}
\usepackage{tikz}
\usetikzlibrary{matrix}
\usepackage{pgf-umlsd}
\usepackage{pgfplots}
\usepackage{pgfplotstable}
\usepgfplotslibrary{fillbetween}
\usepackage{ragged2e}
\usetikzlibrary{arrows,automata,fit,calc,positioning,shapes,shapes.multipart} 
\usepackage{enumerate,url,soul}
\usepackage{paralist}
\usepackage[ruled,vlined]{algorithm2e}
\usepackage{thmtools,thm-restate}
\usepackage{listings}
\usepackage{arydshln}

\usepackage{placeins}
\usepackage{wasysym}
\usepackage{xcolor}

\definecolor{CommentColor}{rgb}{0.90,0.16,0} 

\ifdefined\nocflagdefined

\newcommand{\joerg}[1]{}
\newcommand{\rebecca}[1]{}
\newcommand{\marcel}[1]{}
\newcommand{\alvaro}[1]{}
\newcommand{\todo}[1]{}

\else

\newcommand{\joerg}[1]{\textbf{\color{CommentColor} /* (Joerg) #1  (Joerg) */}}
\newcommand{\rebecca}[1]{\textbf{\color{CommentColor} /* (Rebecca) #1  (Rebecca) */}}
\newcommand{\marcel}[1]{\textbf{\color{CommentColor} /* (Marcel) #1  (Marcel) */}}
\newcommand{\alvaro}[1]{\textbf{\color{CommentColor} /* (Alvaro) #1  (Alvaro) */}}
\newcommand{\todo}[1]{\textbf{\color{CommentColor} /* (TODO) #1  (TODO) */}}

\fi

\newcommand{\citep}[1]{\citeauthor{#1}~\shortcite{#1}}

\definecolor{darkgreen}{HTML}{009900}
\definecolor{lightblue}{HTML}{007acc}

\newcommand{\commentout}[1]{}
\newcommand{\defined}[1]{\textbf{#1}}

\newcolumntype{L}[1]{>{\raggedright\arraybackslash}p{#1}}
\newcolumntype{C}[1]{>{\centering\arraybackslash}p{#1}}
\newcolumntype{R}[1]{>{\raggedleft\arraybackslash}p{#1}}

\newcounter{rowcntr}[table]
\renewcommand{\therowcntr}{(\arabic{rowcntr})}

\newcolumntype{N}{>{\refstepcounter{rowcntr}\therowcntr}c}

\newcommand{\ie}{i.\,e.}

\newcommand{\reals}{{\mathbb{R}}}

\newcommand{\vars}{\ensuremath{V}}
\newcommand{\acts}{\ensuremath{A}}
\newcommand{\init}{\ensuremath{I}}
\newcommand{\goal}{\ensuremath{G}}
\newcommand{\goalhard}{\ensuremath{G^{\text{\textup{hard}}}}}
\newcommand{\goalsoft}{\ensuremath{G^{\text{\textup{soft}}}}}
\newcommand{\cost}{{\ensuremath{c}}}
\newcommand{\pre}{\ensuremath{\mathit{pre}}}
\newcommand{\eff}{\ensuremath{\mathit{eff}}}

\newcommand{\apply}[1]{\ensuremath{[[#1]]}}

\newcommand{\costbound}{{\ensuremath{b}}}

\newcommand{\task}{\ensuremath{\tau}}
\newcommand{\plan}{\ensuremath{\pi}}
\newcommand{\plans}{\ensuremath{\Pi}}

\newcommand{\true}{\ensuremath{\mathit{true}}}

\newcommand{\modelsof}[2]{\ensuremath{{\cal M}_#1(#2)}}

\newcommand{\eifler}{Eif20}

 \usepackage{subcaption}

\title{Iterative Planning with Plan-Space Explanations: A Tool and User Study}

\author{
    \Large \textbf{Rebecca Eifler \and J\"org Hoffmann}\\
    Saarland University, Saarland Informatics Campus, Germany\\
    \{eifler, hoffmann\}@cs.uni-saarland.de
}

\begin{document}

\maketitle

\begin{abstract}
In a variety of application settings, the user preference for a
planning task -- the precise optimization objective -- is difficult to
elicit. One possible remedy is planning as an iterative process,
allowing the user to iteratively refine and modify example plans. A
key step to support such a process are explanations, answering user
questions about the current plan. In particular, a relevant kind of
question is ``Why does the plan you suggest not satisfy $p$?'', where
$p$ is a \emph{plan property} desirable to the user. Note that such a
question pertains to plan space, \ie, the set of possible alternative
plans. Adopting the recent approach to answer such questions in terms
of \emph{plan-property dependencies}, here we implement a tool and
user interface for human-guided iterative planning including
plan-space explanations. The tool runs in standard Web browsers, and
provides simple user interfaces for both developers and users. We
conduct a first user study, whose outcome indicates the usefulness of
plan-property dependency explanations in iterative planning.
\end{abstract}

\section{Introduction}
\label{sec::introduction}

In many real life settings, like space mission control, production
planning in Industry 4.0, or robot-aided disaster recovery, typically
not all goals, constraints and preferences are known from the
beginning. There may even be different user groups with different
preferences. Take for instance researchers from different fields in a
mission control center, who all have to be satisfied with the plan.
Given this setting, the traditional planning workflow -- select a set
of goals, compute a plan, execute -- is not adequate. Instead, planning
should support the users in making up their minds, exploring plan
space until they are satisfied. 

\begin{figure}[ht]
  \centering
  \scriptsize
  \begin{tikzpicture}
    
    \node[draw, minimum width=1.5cm, minimum height=1cm, align=center] (planner) at (0,0) {Planner};
    \node[draw, minimum width=1.5cm, minimum height=1cm, align=center] (plan) at (0,-2.5) {Plan(s)};
    \node[draw, minimum width=1.5cm, minimum height=1cm] (user) at (5,0) {User};

    \draw[->, thick] (user) to 
      node[above, align=center] {Goals\\ Constraints\\ Preferences} 
      node[below, align=center, red] {Plan Properties} 
    (planner);
    \draw[->, thick] (planner) to
      node[left, align=center] {Multiple\\ Qual. distinct\\ Plans}
      node[right, align=center, red] {Plan}
    (plan);
    \draw[->, thick] (plan.east) to
      node[above, align=center, rotate=25, anchor=south] {Explanation}
      node[below, align=center, rotate=25, anchor=north, red] {Plan Property Dependencies}
    (user.south);
    
  \end{tikzpicture}
  \caption{ Planning viewed as an iterative process of plan revision
    under constraints.  black: generic description by Smith
    \shortcite{smith:aaai-12}; red: our instantiation.}
  \label{fig:iter-planning}
\end{figure}
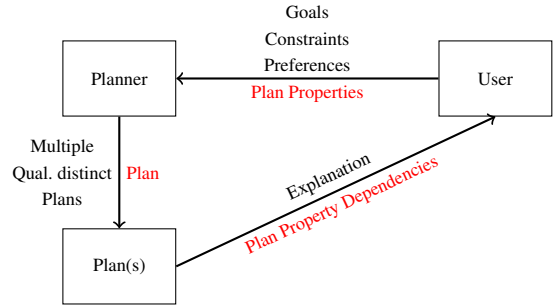

A natural framework for the latter is
\emph{planning as an iterative process} \cite{smith:aaai-12}, as
illustrated in Figure \ref{fig:iter-planning}.
This allows the human users to iteratively refine their preferences
and goals based on example plans.  The possibility to provide
explanations, answering user questions about the current plan, is a
key feature in this setting. In particular, user questions of the form
``Why does the suggested plan $\pi$ not satisfy preference $p$?''  are
relevant. Such explanations allow the user to develop a deeper
understanding of the space of plans. This in turn enables the user to
refine his goals and preferences accordingly.

Eifler et al.\ \shortcite{eifler:etal:aaai-20,eifler:etal:ijcai-20},
henceforth referred to as \emph{\eifler}, introduced a framework
addressing the problem of generating explanations via
\defined{plan-property entailments}.
A plan property $p$ is a Boolean function on plans, and $p$ entails
$q$ if all plans that satisfy $p$ also satisfy $q$. 
The above stated question could then be answered with 
``Because if you achieve $p$ you have to forego $p'$'' .
This is a form of contrastive explanation \cite{miller:ai-19}.  Some
other works
\cite{smith:aaai-12,fox:etal:ijcai-ws-17,cashmore:etal:xaip-19,krarup:etal:xaip-19}
propose to answer such questions by highlighting the differences
between a plan $\pi$ not satisfying $p$ and an alternative plan $\pi'$
that does.
Observe that both approaches naturally shed light on the trade-off
between different user preferences.

\eifler\ assume that the set $P$ of plan properties is given as an
input.  They then address oversubscription planning (OSP)
\cite{smith:icaps-04,domshlak:mirkis:jair-15} where not all of $P$ can
be satisfied.  Based on \defined{exclusion dependencies} of the form
$\bigwedge_{p \in X} p \Rightarrow \neg \bigwedge_{p \in Y} p$ where
$X, Y \subseteq P$, they are able to answer the type of questions
stated above. They also propose an \emph{online} version which allows
to adapt the set $P$ during planning.
Arguably, this framework perfectly fits the iterative planning
workflow described by Smith \shortcite{smith:aaai-12}.  The goals,
constraints and preferences can be expressed as plan properties.
Given this representation, the explanations are then naturally given
as plan property dependencies.

Here we contribute a Web-based iterative planning tool, realizing the
workflow suggested by \cite{smith:aaai-12} and instantiating the
explanation part with \eifler's framework.  The tool enables users to
perform iterative planning with plan properties representing goals and
preferences. One can enforce selected plan properties reflecting
changing preferences across planning iterations, one can ask questions
about the iteratively refined plans, and one can add new plan
properties to hone in on new issues that become apparent during the
iterative planning process. To accommodate layperson users unfamiliar
with planning, our tool comprises a simplified version, with a fixed
set of plan properties, and with an enriched visualization of the
planning task.
The tool runs in standard Web browsers, and provides simple user
interfaces for both, case-study developers and users.

We furthermore conduct a first user study evaluating our tool and
therewith providing the first user-based evaluation of \eifler's
framework. Specifically, we evaluate the usefulness of \eifler's
explanation facilities, by comparing the use of our tool with
vs.\ without these facilities. While the user study is small and
preliminary, its outcome is encouraging, indicating the usefulness of
plan-property dependency explanations in iterative planning.

Section~\ref{sec::background} gives some background on the planning
formalism and \eifler's framework.  Section~\ref{sec:tool-description}
describes the tool implementation.  The preliminary user study is
described and evaluated in Section~\ref{sec:user-study}.
Section~\ref{sec::conclusion} lists work we want to address in the
future and concludes the paper.

\section{Background}
\label{sec::background}

\subsection{Oversubscription Planning}
\label{sec::background:planning}

Our framework is based on a finite-domain variable 
variant of oversubscription planning (OSP)
\cite{smith:icaps-04,domshlak:mirkis:jair-15}.
An \defined{OSP task} is a tuple $\task =
(\vars,\allowbreak\acts,\allowbreak\cost,\allowbreak\init,\allowbreak\goalhard,\allowbreak\goalsoft,\allowbreak\costbound)$
where \vars\ is the set of \defined{variables}, \acts\ is the set of
\defined{actions}, $\cost: \acts \to \reals^+_0$ is the action
\defined{cost} function, \init\ is the \defined{initial state},
\goalhard\ (\goalsoft) is the \defined{hard} (\defined{soft})
\defined{goal}, and $\costbound \in \reals^+_0$ is the \defined{cost
  bound}. A \defined{state} is a complete assignment to $\vars$;
\goalhard\ and \goalsoft\ are partial assignments to \vars, defined on
disjoint sets of variables; each action $a \in \acts$ has a
\defined{precondition} $\pre_a$ and an \defined{effect} $\eff_a$, both
partial assignments to \vars. We refer to variable-value pairs $v=d$
as \defined{facts}, and we identify partial variable assignments with
sets of facts.
An action $a$ is \defined{applicable} in a state $s$ if $\pre_a
\subseteq s$. The outcome state $s\apply{a}$ is the same as $s$ except that
$s\apply{a}(v) = \eff_a(v)$ for those $v$ on which $\eff_a$ is
defined. The outcome state of an iteratively applicable action
sequence $\plan$ is denoted by $s\apply{\plan}$.
A \defined{plan} is an action sequence \plan\ whose summed-up cost is
$\leq \costbound$ and where $\goalhard \subseteq \init\apply{\plan}$.

We follow \eifler\ in not defining a plan utility over
$\goalsoft$. Instead, $\goalsoft$ is a set of plan properties --
including more general plan properties compiled into goal facts -- and
the analysis we provide identifies dependencies between these plan
properties.
The underlying assumption is that the user's preferences
over \goalsoft\ are difficult to elicitate, and not expressable 
in a utility function.

\begin{figure}[ht]
  \centering
  \begin{tikzpicture}
    \node[] (b) at (0,0) {\includegraphics[scale=0.3]{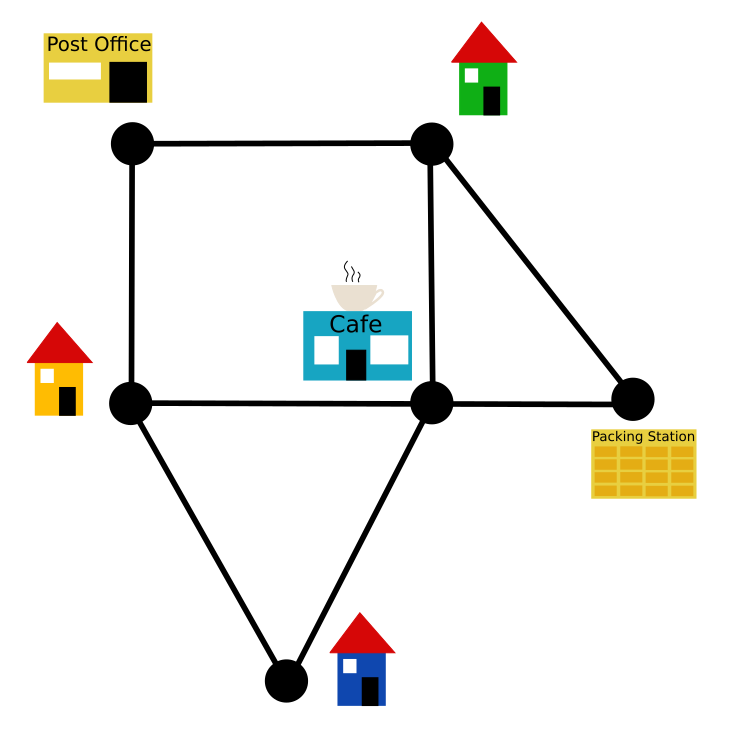}};
    \node[] (rt) at (0.5,-0.2) {\includegraphics[scale=0.1]{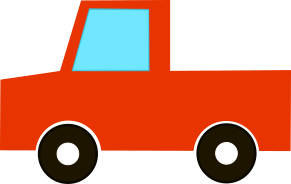}};
    \node[] (rt) at (-1.9,1.9) {\includegraphics[scale=0.1]{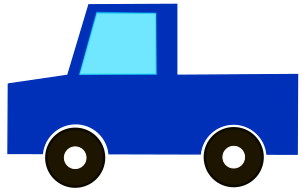}};

    \node[] (p1) at (2.0,-0.9) {\includegraphics[scale=0.06]{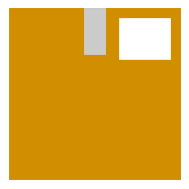}};
    \node[] (p1) at (2.3,-0.9) {0};

    \node[] (p1) at (2.6,-0.9) {\includegraphics[scale=0.06]{package.png}};
    \node[] (p1) at (2.9,-0.9) {1};

    \node[] (p1) at (-3.2,2.2) {\includegraphics[scale=0.06]{package.png}};
    \node[] (p1) at (-2.9,2.2) {2};

    \node[] (p1) at (-2.6,2.2) {\includegraphics[scale=0.06]{package.png}};
    \node[] (p1) at (-2.3,2.2) {3};

    \node[] (p1) at (-2.8,-0.5) {\includegraphics[scale=0.06]{package.png}};
    \node[] (p1) at (-2.5,-0.5) {4};

  \end{tikzpicture}
  
	\caption{An illustrative NoMystery example.}
    \label{fig::nomystery_example}

\end{figure}

As our running example, consider the OSP task illustrated in
Figure~\ref{fig::nomystery_example}, based on the IPC NoMystery domain
where packages must be delivered respecting limited fuel.  The figure
depicts the initial locations of trucks and packages.  Both trucks
have initially 3 fuel units and each road section consumes 1 unit of
fuel. This is also the example we use in our preliminary user study.

\subsection{Eifler et al.'s Framework}
\label{sec::background:framework}

\eifler\ introduce a
framework for plan-property dependency analysis in OSP tasks, and
algorithms for a specific instantiation of that framework. 
For the work presented in the following, we only consider that instantiation, simplifying \eifler's concepts accordingly. 
In the next section, we outline the definitions that build the foundation of our contribution.

Assume an OSP task $\task =
(\vars,\allowbreak\acts,\allowbreak\cost,\allowbreak\init,\allowbreak\goalhard,\allowbreak\goalsoft,\allowbreak\costbound)$. 
In general, a \defined{plan property} is a function $p$ mapping action sequences \plan\ in \task\ to Boolean values. 
\eifler's implementation only, considers plan properties $p$ that can be compiled into goal facts. 
A plan \plan\ satisfies a property $p$ iff the corresponding goal fact $g_p$ is true in $\init\apply{\plan}$. 
They are interested in conjunctions $\bigwedge_{g \in X} g$ of such plan properties, and specifically in
\defined{exclusion dependencies} where $\bigwedge_{g \in X} g$ ``entails'' $\neg \bigwedge_{g \in Y} g ~~ (X,Y \subset P)$.

Entailment here is intended as entailment \emph{in the space of plans}. 
It is often the case that all \emph{plans} which satisfy $X$ cannot satisfy $Y$. 
We denote by \plans\ the set of plans for \task. 
We say that $\plan \in \plans$ \defined{satisfies} a formula $\phi$ over \goalsoft, written $\plan \models \phi$, if $\phi$
evaluates to true under the truth value assignment where $g \in \goalsoft$ is \true\ iff $g \in \init\apply{\plan}$. 
We denote by $\modelsof{\plans}{\phi} := \{\plan \mid \plan \in \plans, \plan \models \phi\}$ the subset of plans that satisfy $\phi$. 
We say that $\phi$ \defined{\plans-entails} $\psi$ if $\modelsof{\plans}{\phi} \subseteq \modelsof{\plans}{\psi}$.

Applied to exclusion dependencies, this means that $\bigwedge_{g \in X} g$ \plans-entails $\neg \bigwedge_{g \in Y} g$ if all action
sequences in \task\ whose cost is within the bound \costbound, that achieve \goalhard, and that achieve all $g \in X$, do not achieve at least one $g \in Y$. 

The problem now is to compute all exclusion dependencies over \goalsoft. 
This corresponds to preparing answers to all questions of the form ``Why does the plan not satisfy the properties in $X$?''.

To this end, observe that $\bigwedge_{g \in X} g$ \plans-entails $\neg
\bigwedge_{g \in Y} g$ iff $\bigwedge_{g \in X \cup Y} g$ is
unsolvable in \task. Observe further that, in this case, the same is
true for every $X' \supseteq X$ and $Y' \supseteq Y$, \ie, the
exclusion dependency is strongest for set-inclusion minimal $X$ and
$Y$. Given these observations, the problem boils down to computing all
\defined{minimal unsolvable goal subsets (MUGS)}: all sets $\goal
\subseteq \goalsoft$ where $\goal$ cannot be achieved but every
$\goal' \subsetneq \goal$ can. 

The question ``Why is the task not solvable?'' can be seen as a special case of goal exclusion 
$\bigwedge_{g \in X} g$ entails $\neg \bigwedge_{g \in Y} g$, where $X$ is empty. 
In general the answer is constituted of all MUGS $\subseteq \goalhard$.
In iterative planning setting an explanation focusing on the latest 
changes leading to unsolvability, allows the user to precisely evaluate his decision. 
Those are the MUGS $\subseteq \goalhard_{i}$ which
contain at least one plan property $p \in \goalhard_{i} \setminus \goalhard_{i-1}$,
where $\goalhard_{i}$ are the hard goals in iteration $i$.

For our example we consider the following plan properties:
\begin{enumerate}
  \label{list:plan_properties}
  \item Package $0$ is delivered to the blue house.
  \item Package $1$ is delivered to the green house.
  \item Package $2$ is delivered to the blue house.
  \item Package $3$ is delivered to the orange house.
  \item Package $4$ is delivered to the post office.
  \item Package $4$ is delivered to the packing station.
  \item The road between the cafe and the packing station is not used by the red truck.
  \item The same truck is used for package 0 and 2.
  \item The same truck is used for package 2 and 3.
  \item The blue truck visits the cafe.
  \item Package 0 is delivered before package 1.
\end{enumerate}
The first six of these are standard goal facts; 7 to 11 can be
expressed in LTL, which can be compiled into goal facts using
well-known techniques \cite{baier:mcilraith:aaai-06,edelkamp:icaps-06,eifler:etal:ijcai-20}; 7
to 10 can also be expressed in a simple LTL fragment named
\emph{action set properties} by \eifler, for which a very compact
compilation into goal facts is possible (and that \eifler\ use in
their implementation whenever possible).

The MUGS in this example are $
\{2, 4, 9\},\allowbreak
\{2, 6\},\allowbreak
\{3, 6, 9\},\allowbreak
\{3, 8\},\allowbreak
\{4, 6, 9, 10\},\allowbreak
\{4, 7, 9\},\allowbreak
\{5, 6\},\allowbreak
\{5, 7\},\allowbreak
\{6, 8, 10\},\allowbreak
\{7, 8\}\allowbreak
$. 
So, for example, the answer to the user question "Why don't you avoid driving from the packing station to the cafe?",
could be answered using MUGS $\{7, 8\}$ "Because then you can not deliver package 0 and 2 with the same truck.".

\section{Tool Description}
\label{sec:tool-description}

\begin{figure*}[ht]
    \centering
    \includegraphics[scale=0.19]{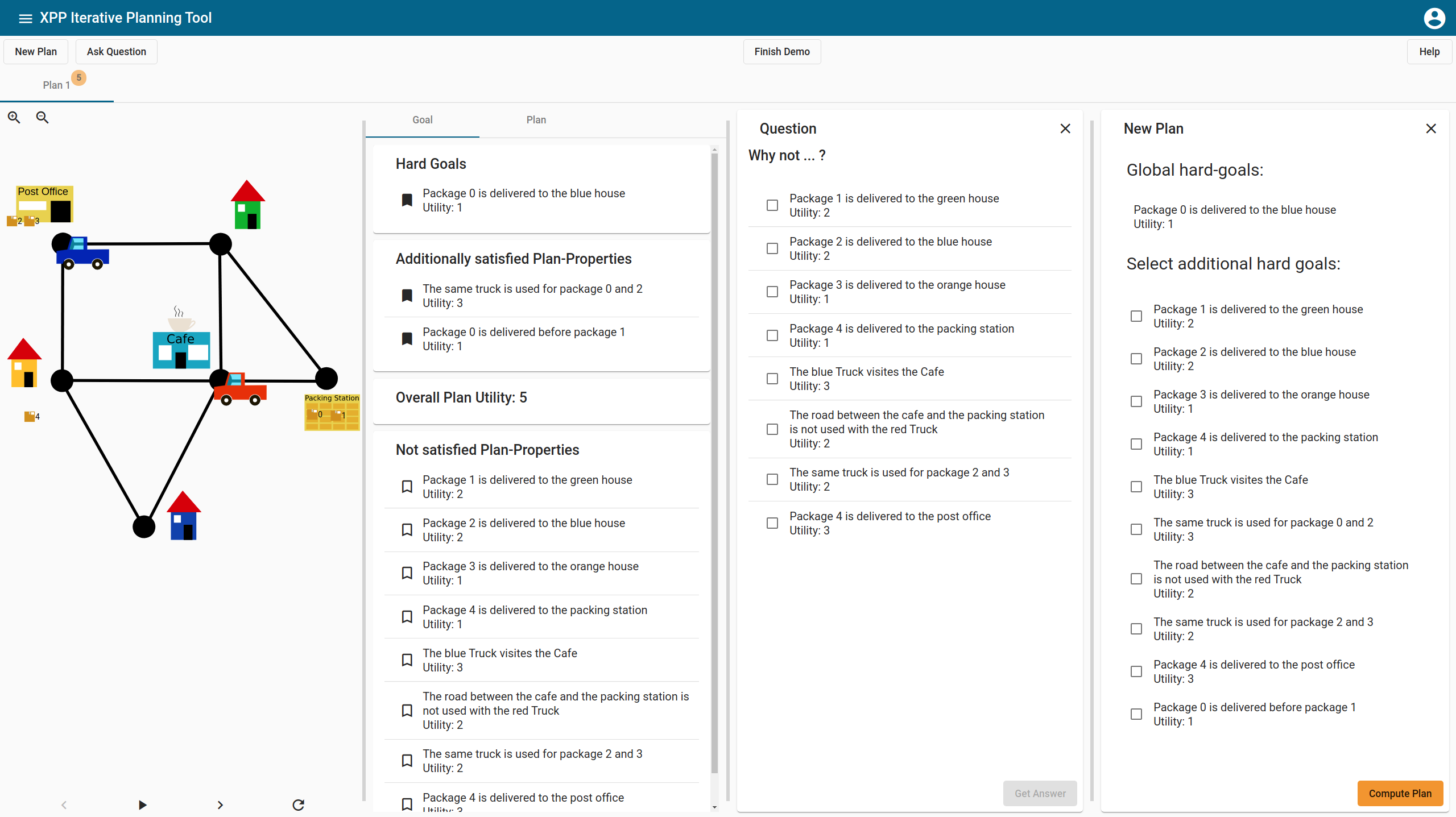}
    \caption{Tool overview: Iterative planning.}
    \label{fig:tool-overview-demo-plan-questions}
\end{figure*}

We implemented a Web based iterative planning tool.
As a planner we use Fast Downward \cite{helmert:jair-06} and as an explanation generator 
the implementation of \eifler.
The translation of plan properties to goal facts is realized using \eifler's definitions 
and implementations for LTL and action-set plan properties.
A demonstration of the platform can be found here \url{https://youtu.be/M958F14HL8I}.

\paragraph{Users}
The tool is designed for two different user groups.  Technically
versed users, familiar with computer science or AI planning; and
layperson users, with no computer science background.  
The adaptations for layperson users focus mainly on the visualization 
of the planning task and plan properties.  
The part of the tool dedicated to layperson users is designed 
to give the possibility to conduct user studies.

\subsection{Overview}
First, we want to give a general overview of the tool structure
and the provided functionalities.
The tool has three main components,
\emph{projects}, \emph{demos} and \emph{user studies}.

A project consists of a planning task definition and plan properties
reflecting the goals and user preferences.
It provides the full functionality for iterative planning with 
online computed explanations.
This includes:
\begin{itemize}
    \item enforcing selected plan properties reflecting changing preferences across planning iterations
    \item asking questions about the iteratively refined plans
    \item adding new plan properties to hone in on new issues that become apparent during the
    iterative planning process
\end{itemize}

A demo provides iterative planning with a fixed set of 
plan properties as a simplified version 
to accommodate layperson users unfamiliar with planning.
To provide a graphical visualization of the planning task and plan
you can add a picture and a domain dependent animation respectively.

The platform supports conducting user studies with the online 
recruitment platform Prolific~\cite{palan:schitter:prolific}.
For user study developers it provides interfaces for setting up and 
evaluating an user study.
A user study can be composed of multiple demos, links to external questionnaires
and additional descriptions and instructions.

\subsection{Input}
Except for the visualization extensions for layer persons, the implementation
is domain independent.

\paragraph{Planning Task} 
The task definition is given by a standard PDDL \cite{pddl-handbook} domain and problem file.
The problem file only defines the initial state. 
The definition of goals is handled separately using plan properties.
It is not possible to modify the initial state or the action definitions during the 
iterative planning process.

\paragraph{Plan Properties}
The plan properties reflect the goals, constraints and user preferences.
They have to be defined by a user.
The platform offers two ways to define plan properties.
Firstly, users familiar with LTLf can define plan properties by explicitly stating the corresponding LTLf formula and action sets.

\begin{figure}[ht]
    \centering
    \includegraphics[scale=0.3]{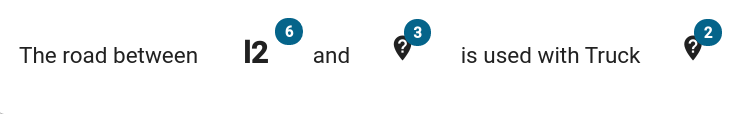}
    \caption{Interface plan property creation.}
    \label{fig:plan-property-crator}
\end{figure}

Secondly, a domain expert can define domain dependent templates for plan properties.
These have to be provided with the domain and problem file.
A template maps a predefined natural language representation to the corresponding LTLf formula. 
For example, the interface for the plan property template "The road between $L_i$ and $L_j$ is used with truck $T_i$"
is given in Figure \ref{fig:plan-property-crator}.
For every variable the type can be specified.
Furthermore, it is possible to impose constraints between the variables, 
like $L_i$ and $L_j$ have to be connected.
This allows to define plan properties more easily as one only has to select the
concrete instantiation. 

Independent of their meaning, e.g. goals, constraints, preferences,
all plan properties are handled the same in the computation of plans
and explanations.
Therefore, the interface does not differentiates between different 
types of plan properties.

Initially, plan properties representing the already known goals and constraints
have to be provided by the user.
As these might change or new preferences might form during planning, it is possible 
to add more plan properties reflecting these changes during planning.

A plan property can be declared as a global hard goal. 
Such plan properties are preselected as hard goals for every planning iteration.
In our example the delivery of package 0 is a global hard goal representing a priority package.

\subsection{Iterative Planning}
An overview of the iterative planning interface is given by
Figure~\ref{fig:tool-overview-demo-plan-questions}.
It is divided into four columns.
The first column contains the graphical representation of the planning task.
The hard goals and the additionally satisfied soft goals of the selected plan are listed in 
the second column.
Explanations for the current plan are provided in the third column.
The fourth column hosts the interface for selecting the hard goals for the next iteration. 
This interface is used by end users and test persons for iterative planning.
The different interface parts will be explained in more detail in the next sections.

\paragraph{Plan Step}

\begin{figure}[ht]
    \centering
    \includegraphics[scale=0.3]{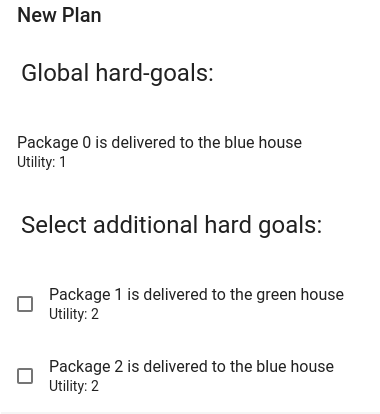}
    \caption{Interface to select plan properties as hard goals.}
    \label{fig:select-pp-plan}
\end{figure}

To compute a plan, the user has to select the hard goals.
The global hard goals are automatically added. 
All not selected plan properties are treated as soft goals.
The corresponding interface for the example task is shown in Figure \ref{fig:select-pp-plan}.
At the top the global hard goals are listed.
The plan properties which can be selected by the user as additional hard goals are enumerated below.
After confirming the choice, a plan for the selection is computed.

For a solvable selection of hard goals, the tool provides the following result:

\begin{itemize}
    \item the selected hard goals
    \item the soft goals which are additionally satisfied although not enforced
    \item the unsatisfied soft goals
    \item the plan as a sequence of actions
    \item the plan as an animation
\end{itemize}

\begin{figure}[ht]
    \centering
    \includegraphics[scale=0.3]{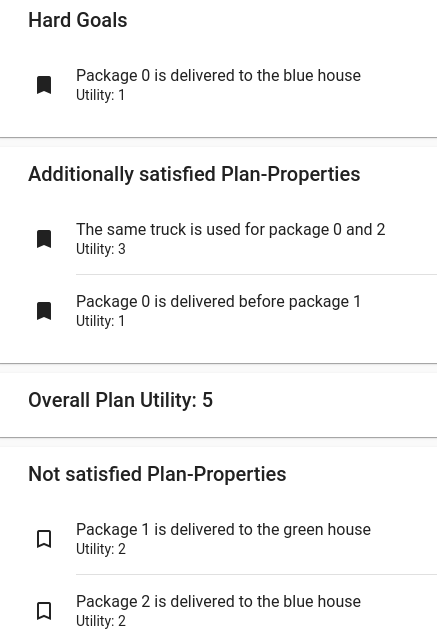}
    \caption{Information about the computed plan.}
    \label{fig:plan-result}
\end{figure}

Figure~\ref{fig:plan-result} depicts the representation of a plan for which no other 
hard goals than the global hard goals have been selected.
In this instance two additional soft goals are satisfied although not enforced. 

If the task is unsolvable, this is explicitly stated to the user and only the hard goals are depicted.

\paragraph{Explanation}

\begin{figure}[ht]
    \centering
    \includegraphics[scale=0.3]{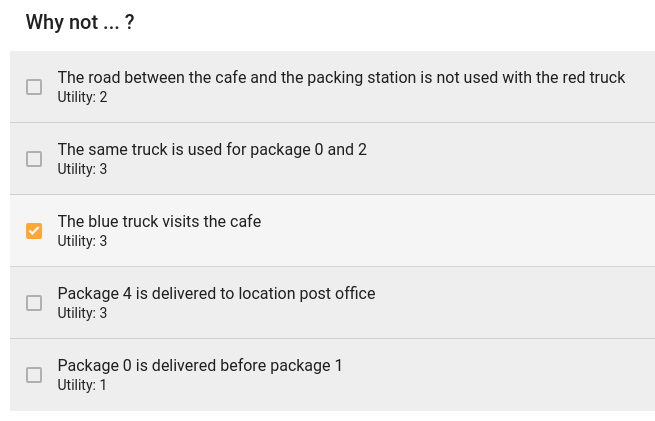}
    \caption[]{Interface to ask a question.}
    \label{fig:ask_question}
\end{figure}

In order to get more insights into the space of plan, the user can ask questions about the current plan.
More specifically, he can select a subset $P_N$ of the plan properties not satisfied by the current plan.
The selection is then interpreted as the question: "Why are the plan properties in $P_N$ not satisfied?".
The interface to ask a question is depicted in Figure~\ref{fig:ask_question}.
The resulting question would be: "Why does the blue truck not visit the cafe?"

\begin{figure}[ht]
    \centering
    \includegraphics[scale=0.3]{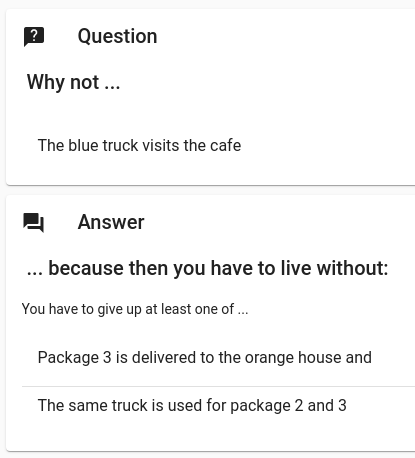}
    \caption[]{Answer to the question of Figure~\ref{fig:ask_question}.}
    \label{fig:get_answer}
\end{figure}

The answer is then provided as shown in Figure~\ref{fig:get_answer}.
It is a list of all MUGS containing one of the plan properties in the question.
Within the answer, only plan properties which are currently satisfied by the plan (enforced or by chance) are considered.
The answer depicted in Figure~\ref{fig:get_answer} can be summarized as: 
"If the blue truck visits the cafe, then either package 3 can not be delivered 
or package 2 and 3 are not transported with the same truck."

If a selection of hard goals is not solvable, the user can ask ``Why?''.
The answer is given as all minimal unsolvable subsets of the hard goals.

\subsection{User Study Support}
To allow a layperson user to understand the planning task, the plan properties and the
generated plans more easily, we implemented the following adaptations.
The initial state of the planning task is visualized with an image. 
The plan actions can be substituted with domain dependent natural language sentences.
Additionally, a domain dependent animation for each plan can be added. 
Also the natural language representation of plan properties can be task dependent.
In the example this allows to take the map of the task into account, instead of 
using generic location names.

To conduct a user study with a feasible amount of delay and robustness
there exists the possibility to compute a \emph{demo} from a project.
To do so, the available plan properties have to be fixed and all MUGS are precomputed.
To further reduce delay it could also be possible to precompute all plans.
Due to the large number of combinations of hard goals, this approach might
oftentimes not be feasible and is therefore not implemented by the tool.

Given such a demo, the iterative planning workflow can be further restricted for a 
user study:
The possibility to ask questions can be enabled or disabled.
Furthermore, the amount of time and/or the number of iterations available to the user can
be restricted.
In addition, the size of the question a user can ask can be limited.

If iterative planning is performed in the setting of a user study, the time a test person 
spends with each interface part is measured. This includes for example, how much time she spends 
with asking questions or selecting the hard goals for the next iteration.

\section{User Study}
\label{sec:user-study}

We evaluated the usefulness of the explanations provided by the framework of \eifler\
in the setting of iterative planning in a small preliminary user study.

\subsection{User Preferences}
The main purpose of the tool and the explanations is to support a user 
in the process of finding and understanding a plan which satisfies 
\emph{her preferences}.
In many domains only an expert user in the field has such intrinsic preferences.
But such test persons are very difficult to recruit.
To conduct a reasonable user study in a crowd sourcing setting, it is necessary
to provide artificial preferences to the user. 
Otherwise, the lack of own user preferences leads to meaningless plan properties 
and plan differences for the user.
Therefore, we decided to assign a utility as an integer value to each plan property.
This is actually a contradiction to our assumption, that user preferences are difficult
to elicitat, and not expressable in a utility function.
But it is essential for a successful user study, as it provides the following two features.
First of all, it gives the user different preferences for the different plan properties.
Additionally, it is possible to motivate the user to find good plans, reflected by a high utility. 
This can be achieved by rewarding the user based on the best plan utility he was able
to accomplish.
In order to make use of this feature, the utility of each plan property 
in association with its natural language description, is shown to the user as depicted in Figure~\ref{fig:ask_question}.

\subsection{Setup}
The test persons were 6 planning researchers from the FAI Group of Saarland University.
They were all familiar with the framework, but not with the example and 
they used the tool for the first time. 
The user study was split into three parts.
First, the test persons were provided with a manual of the tool and had time 
to understand the functionality provided by it. 
Then, they could familiarize themselves with the planning task in Figure~\ref{fig::nomystery_example},
and the plan properties listed in Section~\ref{sec::background}.
The delivery of package 0 was fixed as a global hard goal.

In the second part, each plan property was assigned a utility between 1 and 3.
All test persons were asked to maximize the overall utility of the plan 
properties satisfied by a plan.
The maximal possible utility was 14.
Again, this is solely done in order to provide some extrinsic motivation.

To be able to evaluate the \emph{usefulness} of the explanations, the test persons
are split into two groups of size 3 each.
Group GQ+ was allowed to ask questions, group GQ- was not.
The maximum number of iterations was fixed to 10 and the maximal question size to 1.
The test persons were not able to ask questions about an unsolvable task.
\footnote{This is only due to the fact that this feature had not been 
yet implemented at the time the study was conducted.}

In the third part, the test persons who could ask questions had to answer a small
questionnaire with respect to the helpfulness of the provided explanations.

The questionnaire was constituted of the following questions:
\begin{enumerate}
    \item How helpful were the provided explanations in the setting of iterative planning?
    \item How helpful were the provided explanations to understand the space of plans?
\end{enumerate}

The questions could be answered using a discrete linear scale reaching
from 1 (Didn't help at all) to 5 (Helped a lot).  Additionally, there
was a free text question asking for any feedback with respect to the
provided explanations and the implementation of the tool itself.

\subsection{Evaluation}

Given the small number of test persons the resulting quantitative
evaluation has to be interpreted with care.  Nevertheless, the data
can give some idea of the general tendency.

\paragraph{Numeric Evaluation}

\begin{figure}[ht]
    \centering
\small
\begin{tikzpicture}
    \begin{axis}[
        title=GQ-,
        height=4cm,
        width=8cm,
        xlabel=iteration,
        y label style={at={(axis description cs:+0.15,.5)},anchor=south},
        ylabel=utility,
        xtick = {0,1,2,3,4,5,6,7,8,9},
        ytick = {0,2,4,6,8},
        yticklabels={0,2,4,6,8},
        ymin =-0.5,
        xmin=-0.5,
        xmax=9.5,
        ybar,
        bar width=0.1cm
    ]
        \addplot+ [blue] coordinates{
            (0,5) (1,0) (2,0) (3,0) (4,0)
        };
        \addplot+ [red] coordinates{
            (0,0) (1,0) (2,8) (3,0) (4,0) (5,6) (6,9) (7,7) (8,8) (9,9)
        };
        \addplot+ [darkgreen] coordinates{
            (0,0) (1,0) (2,0) (3,0) (4,0) (5,8) (6,8) (7,0) (8,0) (9,0)
        };
    \end{axis}
    \draw[thick, blue, dashed] (3.2,2.4) to (3.2,0);
    \draw[thick, red, dashed] (6.3,2.4) to (6.3,0);
    \draw[thick, darkgreen, dashed] (6.25,2.4) to (6.25,0);

    \node[red] (t1) at (0.35,0.3) {\Lightning};
    \node[darkgreen] (t1) at (0.5,0.3) {\Lightning};

    \node[blue] (t1) at (0.85,0.3) {\Lightning};
    \node[red] (t1) at (1.0,0.3) {\Lightning};
    \node[darkgreen] (t1) at (1.15,0.3) {\Lightning};

    \node[blue] (t1) at (1.45,0.3) {\Lightning};
    \node[darkgreen] (t1) at (1.80,0.3) {\Lightning};

    \node[blue] (t1) at (2.1,0.3) {\Lightning};
    \node[red] (t1) at (2.25,0.3) {\Lightning};
    \node[darkgreen] (t1) at (2.45,0.3) {\Lightning};

    \node[blue] (t1) at (2.75,0.3) {\Lightning};
    \node[red] (t1) at (2.90,0.3) {\Lightning};
    \node[darkgreen] (t1) at (3.05,0.3) {\Lightning};

    \node[darkgreen] (t1) at (5.0,0.3) {\Lightning};

    \node[darkgreen] (t1) at (5.65,0.3) {\Lightning};

\end{tikzpicture}
    
\begin{tikzpicture}
    \node[] (space) at (0,3.5) {};
    \begin{axis}[
        title=GQ+,
        height=4cm,
        width=8cm,
        y label style={at={(axis description cs:+0.15,.5)},anchor=south},
        ylabel=utility,
        xtick = {0,1,2,3,4,5,6,7,8,9},
        ytick = {0,2,4,6,8},
        yticklabels={0,2,4,6,8},
        ymin =-0.5,
        xmin=-0.5,
        xmax=9.5,
        ybar,
        bar width=0.1cm
    ]
        \addplot+ [blue] coordinates{
            (0,8) (1,8) (2,0) (3,8) (4,9)
        };
        \addplot+ [red] coordinates{
            (0,0) (1,0) (2,9) (3,5) (4,5) (5,0) (6,0) (7,9)
        };
        \addplot+ [darkgreen] coordinates{
            (0,0) (1,0) (2,0) (3,8) (4,9) (5,9)
        };
    \end{axis}
    \draw[thick, blue, dashed] (3.2,2.4) to (3.2,0);
    \draw[thick, red, dashed] (5.1,2.4) to (5.1,0);
    \draw[thick, darkgreen, dashed] (3.9,2.4) to (3.9,0);

    \node[red] (t1) at (0.35,0.3) {\Lightning};
    \node[darkgreen] (t1) at (0.5,0.3) {\Lightning};

    \node[red] (t1) at (1.0,0.3) {\Lightning};
    \node[darkgreen] (t1) at (1.15,0.3) {\Lightning};

    \node[blue] (t1) at (1.45,0.3) {\Lightning};
    \node[darkgreen] (t1) at (1.80,0.3) {\Lightning};

    \node[red] (t1) at (3.55,0.3) {\Lightning};

    \node[red] (t1) at (4.2,0.3) {\Lightning};

\end{tikzpicture}
\begin{tikzpicture}
    \begin{axis}[
        height=4cm,
        width=8cm,
        xlabel=iteration,
        y label style={at={(axis description cs:+0.15,.5)},anchor=south},
        ylabel=\#questions,
        xtick = {0,1,2,3,4,5,6,7,8,9},
        ytick = {0,1,2,3},
        yticklabels={0,1,2,3},
        ymax=3.5,
        ymin =-0.2,
        xmin=-0.5,
        xmax=9.5,
        ybar,
        bar width=0.1cm
    ]
        \addplot+ [blue] coordinates{
            (0,1) (1,3) (3,1) (4,1)
        };
        \addplot+ [red] coordinates{
            (2,2) (3,1) (4,0) (7,0)
            
        };
        \addplot+ [darkgreen] coordinates{
            (3,1) (4,1) (5,1)
        };
    \end{axis}
    \draw[thick, blue, dashed] (3.2,2.4) to (3.2,0);
    \draw[thick, red, dashed] (5.1,2.4) to (5.1,0);
    \draw[thick, darkgreen, dashed] (3.9,2.4) to (3.9,0);

    \node[red] (t1) at (0.35,0.3) {\Lightning};
    \node[darkgreen] (t1) at (0.5,0.3) {\Lightning};

    \node[red] (t1) at (1.0,0.3) {\Lightning};
    \node[darkgreen] (t1) at (1.15,0.3) {\Lightning};

    \node[blue] (t1) at (1.45,0.3) {\Lightning};
    \node[darkgreen] (t1) at (1.80,0.3) {\Lightning};

    \node[red] (t1) at (3.55,0.3) {\Lightning};

    \node[red] (t1) at (4.2,0.3) {\Lightning};

\end{tikzpicture}

     \caption{Plan utility and \#questions per iteration. 
    The different colors correspond to the different test persons.
    Dashed lines indicate max number of used iterations.
    \Lightning\ indicates that the selected hard goals were unsolvable in this iteration.}
    \label{fig:utility-iterations}
\end{figure}
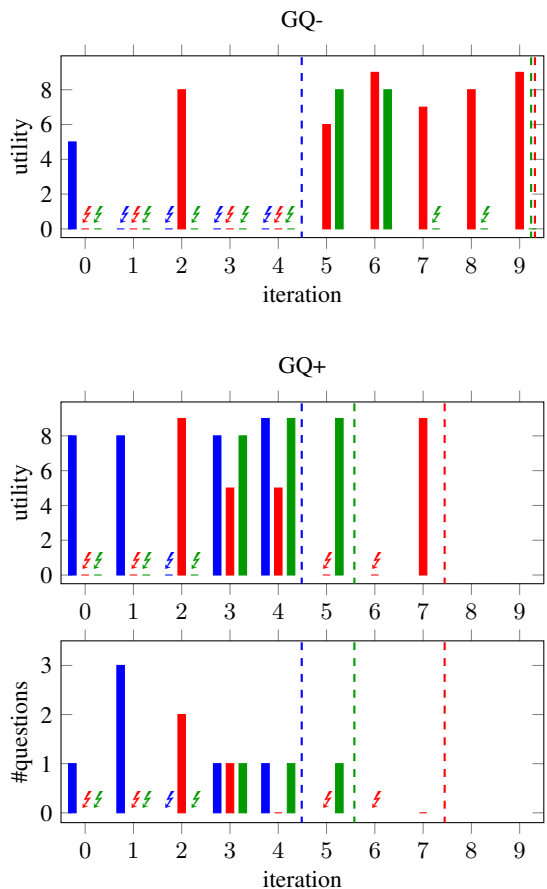

The top two graphics in Figure~\ref{fig:utility-iterations} depict 
the overall utility achieved per iteration
per test persons with and without explanations.
Based on the selection of test persons a statistical analysis is not adequate.
Nevertheless, For the sake of completeness, we give the $95\%$ confidence interval
for every expected value for the utility and the number of iterations.
One of the test persons in GQ- quit after 5 iterations without reaching
a reasonable high utility. 
Therefore, we will evaluate GQ- without the blue test person.
The test persons in GQ-, on average, achieved a plan utility of $8.5$ $[7.69, 9.30]$ (7.3, when 
including the blue test person).
Both test persons in GQ- used all $10$ iterations (when 
including the blue test person, on average $8.3$).
All test persons in GQ+ achieved a maximal plan utility of $9$.
They came up with their final plan within $6.3$ $[4.63, 8.03]$ iterations on average.
The average number of questions asked in total was 4.
The bottom graphic in Figure~\ref{fig:utility-iterations} depicts the 
number of questions per iteration per test person.
Although, the strategy of all test persons varied a lot, 
the questions helped to achieve a higher utility with less iterations.
In addition, being able to ask questions led to a smaller amount of unsolvable tasks.

The results for both questions in the questionnaire are on average $3.66$ with a 
variance of $0.24$ on a scale from 1 (Didn't help at all) to 5 (Helped a lot).

\paragraph{Free Text Feedback}

In the following we evaluate and summarize the feedback given in the free text question.
In general, the test persons found the possibility tho ask questions useful, as it
allowed them to select plan properties more strategically.

In order to expose the functionality provided by the framework in an even more user friendly
way, the interface should be further improved.
For example, all participants agreed, that the interface displaying the answer needs some
improvement.
Especially, in situations where the answer represents multiple MUGS it has to be arranged
more clearly.
Some proposed to use a small guided introduction task to clearly outline the task and how the explanations could help the user.

One test person stated, that the answer should only point out the plan properties 
which are enforced in the plan and not only satisfied by chance.

In this user study the question size was fixed to one although larger questions are possible  in general.
All participants agreed, that in some cases larger questions with a size up to three would be more useful.
In this case, the interface representing the answer would have to clearly indicate the 
influence of the different plan properties in the question.

Almost all test persons started with an unsolvable selection of hard goals.
The questions we currently support are all based on a given plan.
This has the effect that the functionality of asking questions can only be
exploited after a plan for a solvable selection of hard goals is found.
As a consequence, most of the test persons complained, that the possibility of asking 
a question like "Why is this selection of hard goals unsolvable?" was missing.

\section{Conclusion and Future Work}
\label{sec::conclusion}

\eifler\ introduced a framework using plan properties and 
their dependencies to compute plan space explanations.
We implemented a Web tool using this framework in the setting of 
iterative planning as proposed by \cite{smith:aaai-12}.
Additionally, we conducted a small user study showing that in general the 
provided questions are useful in the setting of iterative planning.

Our next goal is to run a user study with more test persons outside the community.
One of the challenges will be the selection of domains, that are suitable for the 
target audience of test persons.
Besides the logistics domain we plan to use a mars rover mission control domain as 
another example from the planning benchmarks. 
Ideally, a test person has its own preferences with respect to the domain.
Consider for example students, which have to schedule their courses for the next term 
as used by \cite{grover:etal:hui20}.
As a domain a vast majority of people is familiar with, we intend to use
the task of driving home from work and taking care of some household duties
like shopping or disposing the waste paper.
We will also consider further metrics proposed by \cite{hoffman:etal:arXiv18} to evaluate 
the explanation framework.

One interesting question for the future is the combination with the framework of \cite{krarup:etal:xaip-19}.
Their framework is based on explanations based on examples.
The combination of both frameworks will allow us to evaluate the better explanation 
for specific situations and to combine them accordingly.

\section*{Acknowledgments}

This material is based upon work supported by the Air Force Office of
Scientific Research under award number FA9550-18-1-0245. Rebecca
Eifler was also supported by the German Research Foundation (DFG) as
part of CRC 248 (see perspicuous-computing.science).

\bibliographystyle{aaai.bst}

\begin{thebibliography}{}

\bibitem[\protect\citeauthoryear{Baier and
  McIlraith}{2006}]{baier:mcilraith:aaai-06}
Baier, J.~A., and McIlraith, S.~A.
\newblock 2006.
\newblock Planning with first-order temporally extended goals using heuristic
  search.
\newblock In {\em Proc.\ AAAI},  788--795.

\bibitem[\protect\citeauthoryear{Cashmore \bgroup et al\mbox.\egroup
  }{2019}]{cashmore:etal:xaip-19}
Cashmore, M.; Collins, A.; Krarup, B.; Krivic, S.; Magazzeni, D.; and Smith, D.
\newblock 2019.
\newblock Towards explainable {AI} planning as a service.
\newblock In {\em ICAPS XAIP}.

\bibitem[\protect\citeauthoryear{Domshlak and
  Mirkis}{2015}]{domshlak:mirkis:jair-15}
Domshlak, C., and Mirkis, V.
\newblock 2015.
\newblock Deterministic oversubscription planning as heuristic search:
  Abstractions and reformulations.
\newblock {\em JAIR} 52:97--169.

\bibitem[\protect\citeauthoryear{Edelkamp}{2006}]{edelkamp:icaps-06}
Edelkamp, S.
\newblock 2006.
\newblock On the compilation of plan constraints and preferences.
\newblock In {\em ICAPS},  374--377.

\bibitem[\protect\citeauthoryear{Eifler \bgroup et al\mbox.\egroup
  }{2020a}]{eifler:etal:aaai-20}
Eifler, R.; Cashmore, M.; Hoffmann, J.; Magazzeni, D.; and Steinmetz, M.
\newblock 2020a.
\newblock A new approach to plan-space explanation: Analyzing plan-property
  dependencies in oversubscription planning.
\newblock In {\em AAAI}.

\bibitem[\protect\citeauthoryear{Eifler \bgroup et al\mbox.\egroup
  }{2020b}]{eifler:etal:ijcai-20}
Eifler, R.; Steinmetz, M.; Torralba, A.; and Hoffmann, J.
\newblock 2020b.
\newblock Plan-space explanation via plan-property dependencies: Faster
  algorithms \& more powerful properties.
\newblock In {\em IJCAI},  4091--4097.

\bibitem[\protect\citeauthoryear{Fox, Long, and
  Magazzeni}{2017}]{fox:etal:ijcai-ws-17}
Fox, M.; Long, D.; and Magazzeni, D.
\newblock 2017.
\newblock Explainable planning.
\newblock In {\em ICAI XAI}.

\bibitem[\protect\citeauthoryear{Grover \bgroup et al\mbox.\egroup
  }{2020}]{grover:etal:hui20}
Grover, S.; Sengupta, S.; Chakraborti, T.; Mishra, A.~P.; and Kambhampati, S.
\newblock 2020.
\newblock Radar: automated task planning for proactive decision support.
\newblock {\em Human--Computer Interaction}  1--26.

\bibitem[\protect\citeauthoryear{Helmert}{2006}]{helmert:jair-06}
Helmert, M.
\newblock 2006.
\newblock The {Fast} {Downward} planning system.
\newblock {\em JAIR} 26:191--246.

\bibitem[\protect\citeauthoryear{Hoffman \bgroup et al\mbox.\egroup
  }{2018}]{hoffman:etal:arXiv18}
Hoffman, R.~R.; Mueller, S.~T.; Klein, G.; and Litman, J.
\newblock 2018.
\newblock Metrics for explainable ai: Challenges and prospects.
\newblock {\em arXiv preprint arXiv:1812.04608}.

\bibitem[\protect\citeauthoryear{Krarup \bgroup et al\mbox.\egroup
  }{2019}]{krarup:etal:xaip-19}
Krarup, B.; Cashmore, M.; Magazzeni, D.; and Miller, T.
\newblock 2019.
\newblock Towards model-based contrastive explanations for explainable
  planning.
\newblock In {\em ICAPS XAIP}.

\bibitem[\protect\citeauthoryear{McDermott \bgroup et al\mbox.\egroup
  }{1998}]{pddl-handbook}
McDermott, D.; Ghallab, M.; Howe, A.; Knoblock, C.; Ram, A.; Veloso, M.; Weld,
  D.; and Wilkins, D.
\newblock 1998.
\newblock {\em The {PDDL} Planning Domain Definition Language}.
\newblock The {AIPS-98} Planning Competition Comitee.

\bibitem[\protect\citeauthoryear{Miller}{2019}]{miller:ai-19}
Miller, T.
\newblock 2019.
\newblock Explanation in artificial intelligence: Insights from the social
  sciences.
\newblock {\em AI} 267:1--38.

\bibitem[\protect\citeauthoryear{Palan and
  Schitter}{2018}]{palan:schitter:prolific}
Palan, S., and Schitter, C.
\newblock 2018.
\newblock Prolific.ac — a subject pool for online experiments.
\newblock {\em Journal of Behavioral and Experimental Finance} 17:22--27.

\bibitem[\protect\citeauthoryear{Smith}{2004}]{smith:icaps-04}
Smith, D.~E.
\newblock 2004.
\newblock Choosing objectives in over-subscription planning.
\newblock In {\em ICAPS},  393--401.

\bibitem[\protect\citeauthoryear{Smith}{2012}]{smith:aaai-12}
Smith, D.
\newblock 2012.
\newblock Planning as an iterative process.
\newblock In {\em AAAI},  2180--2185.

\end{thebibliography}

\end{document}